\documentclass[journal]{IEEEtran}
\ifCLASSINFOpdf
\else
\fi

\usepackage{cite}
\usepackage[colorlinks,citecolor=green]{hyperref} 
\usepackage{amssymb}
\usepackage{amsmath}
\usepackage{multirow}
\usepackage{booktabs}
\usepackage{lettrine}
\usepackage{subfigure}
\usepackage[misc]{ifsym} 
\usepackage{bicaption}
\ifCLASSINFOpdf
  \usepackage[pdftex]{graphicx}
  \graphicspath{{../pdf/}{../jpeg/}}
  \DeclareGraphicsExtensions{.pdf,.jpeg,.png}
\else
  \usepackage[dvips]{graphicx}
  \graphicspath{{../eps/}}
  \DeclareGraphicsExtensions{.eps}
\fi

\begin{document}
\captionsetup{font={small}}
%
\title{DS-TransUNet: Dual Swin Transformer U-Net\\ for Medical Image Segmentation}
%
%
%

\author{Ailiang~Lin,
        Bingzhi~Chen,
        Jiayu~Xu,
        Zheng~Zhang,~\IEEEmembership{Senior~Member,~IEEE,}
        and~Guangming~Lu,~\IEEEmembership{Member,~IEEE}
\thanks{Ailiang Lin, Bingzhi Chen, Jiayu Xu and Guangming Lu are with the Shenzhen Medical Biometrics Perception and Analysis Engineering Laboratory, Harbin Institute of Technology, Shenzhen 518055, China. (e-mail: tianbaoge24@gmail.com, chenbingzhi@stu.hit.edu.cn, jiayuxu1998@gmail.com, luguangm@hit.edu.cn)}
\thanks{Zheng Zhang is with the Bio-Computing Research Center, Harbin Institute of Technology, Shenzhen 518055, China, and also with Shenzhen Key
Laboratory of Visual Object Detection and Recognition, Shenzhen 518055,
China.(e-mail: darrenzz219@gmail.com)}
}

%
%

\markboth{Journal of \LaTeX\ Class Files,~Vol.~14, No.~8, June~2021}%
{Shell \MakeLowercase{\textit{et al.}}: Bare Demo of IEEEtran.cls for IEEE Journals}
%



\maketitle

\begin{abstract}
Automatic medical image segmentation has made great progress benefit from the development of deep learning. However, most existing methods are based on convolutional neural networks (CNNs), which fail to build long-range dependencies and global context connections due to the limitation of receptive field in convolution operation. Inspired by the success of Transformer whose self-attention mechanism has the powerful abilities of modeling the long-range contextual information, some researchers have expended considerable efforts in designing the robust variants of Transformer-based U-Net. Moreover, the patch division used in vision transformers usually ignores the pixel-level intrinsic structural features inside each patch. To alleviate these problems, in this paper, we propose a novel deep medical image segmentation framework called Dual Swin Transformer U-Net (DS-TransUNet), which might be the first attempt to concurrently incorporate the advantages of hierarchical Swin Transformer into both encoder and decoder of the standard U-shaped architecture to enhance the semantic segmentation quality of varying medical images. Unlike many prior Transformer-based solutions, the proposed DS-TransUNet first adopts dual-scale encoder subnetworks based on Swin Transformer to extract the coarse and fine-grained feature representations of different semantic scales. As the core component for our DS-TransUNet, a well-designed Transformer Interactive Fusion (TIF) module is proposed to effectively establish global dependencies between features of different scales through the self-attention mechanism, in order to make full use of these obtained multi-scale feature representations. Furthermore, we also introduce the Swin Transformer block into decoder to further explore the long-range contextual information during the up-sampling process. Extensive experiments across four typical tasks for medical image segmentation demonstrate the effectiveness of DS-TransUNet, and show that our approach significantly outperforms the state-of-the-art methods.
\end{abstract}


%
\IEEEpeerreviewmaketitle

\section{Introduction}
%
%
%
%
\IEEEPARstart{M}{edical} image segmentation is an important yet challenging research problem involving many common tasks in clinical applications, such as polyp segmentation, lesion segmentation, cell segmentation, etc. Moreover, medical image segmentation is a complex and key step in the field of medical image processing and analysis, and plays an important role in computer-aided clinical diagnosis system. Its purpose is to segment the parts with special significance in medical images and extract relevant features through semi-automatic or automatic process, so as to provide reliable basis for clinical diagnosis and pathological research, and assist doctors in making more accurate diagnosis.\par
With the development of deep learning, convolutional neural networks (CNNs) have become dominant in a series of medical image segmentation tasks. Among various CNN variants, the typical encoder-decoder based network U-Net\cite{ronneberger2015u} has demonstrated excellent segmentation potential, where encoder extracts features through continuous down-sampling, and then decoder progressively leverage features output from encoder through skip connection for up-sampling, so that the network can obtain features of different granularity for better segmentation. Following the popularity of U-Net, many novel models have been proposed such as UNet++\cite{zhou2018unet++}, Res-UNet\cite{xiao2018weighted}, Attention U-Net\cite{oktay2018attention}, DenseUNet\cite{li2018h}, R2U-Net\cite{alom2018recurrent}, KiU-Net\cite{valanarasu2020kiu} and UNet 3+\cite{huang2020unet}, which are specially designed for medical image segmentation and achieve expressive performance. Although CNNs have made great success in the field of medical image, it is difficult for them to make further breakthroughs. Due to the inherent inductive biases, each convolutional kernel can only focus on a sub-region in the whole image, which makes it lose global context and fail to build long-range dependencies. The stacking of convolution layer and down-sampling helps expand the receptive filed and bring better local interaction, but this is a sub-optimal choice because it makes the model more complicated and easier to overfit. There exists some works trying to model long-range dependencies for convolution such as attention mechanism\cite{wang2018non}\cite{huang2019ccnet}\cite{hou2021coordinate}. However, since these methods are not aimed at the field of medical image segmentation, they still have great limitations in global context modeling which means there is great potential for improvement.\par
Recently, the novel architecture Transformer\cite{vaswani2017attention} which was originally designed for sequence-to-sequence modeling in natural language processing (NLP) tasks, has sparked tremendous discussion in computer vision (CV) community. Transformer can revolutionize most NLP tasks such as machine translation, named-entity recognition and question answering, mainly because multi-head self attention (MSA) mechanism can effectively build global connection between the tokens of sequences. The ability of long-range dependencies modeling is also suitable for pixel-based CV tasks. Specially, DEtection TRansformer (DETR)\cite{carion2020end} utilizes a elegant design based on Transformer to build the first fully end-to-end object detection model. Vision Transformer (ViT)\cite{dosovitskiy2020image}, the first image recognition model purely based on Transformer is proposed and achieves comparable performance with other state-of-the-art (SOTA) convolution-based methods. To reduce the computational complexity, a hierarchical Swin Transformer\cite{liu2021swin} is proposed with Window based MSA (W-MSA) and Shifted Window based MSA (SW-MSA) as illustrated in Fig. \ref{trans2}, and surpasses the previous SOTA methods in image classification, dense prediction tasks such as object detection and semantic segmentation. SEgmentation TRansformer (SETR)\cite{zheng2020rethinking} shows that Transformer can achieve SOTA performance in segmentation tasks as encoder. However, Transformer-based models have not attracted enough attention in medical image segmentation. TransUNet\cite{chen2021transunet} utilizes CNNs to extract features and then feeds them into Transformer for long-range dependencies modeling. TransFuse\cite{zhang2021transfuse} based on ViT tries to fuse the features extracted by Transformer and CNNs, while MedT\cite{valanarasu2021medical} based on Axial-Attention\cite{wang2020axial} explores the feasibility of applying Transformer without large-scale datasets. The success of these models shows the great potential of Transformer in medical image segmentation, but they all only apply Transformer in encoder, which means such potential of Transformer in decoder for segmentation remains to be validated.\par
Moreover, multi-scale feature representations have been proved to play an important role in vision transformers. Cross-Attention Multi-Scale Vision Transformer (CrossViT)\cite{chen2021crossvit} proposes a novel dual-branch Transformer architecture to extract multi-scale features for image classification. Multi Vision Transformers (MViT)\cite{fan2021multiscale} is present for video and image recognition by connecting multi-scale feature hierarchies with transformer models. Multi-modal Multi-scale TRansformer (M2TR)\cite{wang2021m2tr} uses a multi-scale transformer to detect the local inconsistency at different scales. In general, multi-scale feature presentations can bring more powerful performance to vision transformers, but they are rarely used in the filed of image segmentation.\par
To alleviate the inherent inductive biases of CNNs, this paper proposes a novel encoder-decoder Transformer based framework that mainly combines the advantages of Swin Transformer and mul      ti-scale vision transformers to effectively optimize the structure of the standard U-shaped architecture for automatic medical image segmentation. Instead of using the traditional encoder structure, the proposed DS-TransUNet adopts dual-scale encoder subnetworks based on hierarchical Swin Transformer under the different scales of image inputs. Specifically, each medical image is first sliced into non-overlapping patches at large and small scales, respectively. By taking these two different scale patches as inputs,  the proposed dual-scale encoder subnetworks can effectively extract the coarse and fine-grained feature representations of different semantic scales, respectively. To make full use of these obtained features, a robust Transformer Interactive Fusion (TIF) module is designed to aggregate the multi-scale feature representations of Swin Transformer between these two encoder subnetworks, which is the key to our DS-TransUNet method. In particular, the coarse-fine-tuning feature representations from two encoder branches will be reshaped into a token of specified size, and then fed into the TIF module to perform an effective interaction potential with each other through the self-attention mechanism of the standard Transformer. Moreover, we also introduce the Swin Transformer block into the decoder, which helps build long-range dependencies and global context connections during up-sampling. Finally, the fused features are gradually restored to the same resolution as the input images for pixel-level predictions. Benefitting from these improvements, the proposed DS-TransUNet can effectively improve the semantic segmentation quality of medical images. 
We evaluate the effectiveness of DS-TransUNet across four typical tasks of medical image segmentation, covering the datasets of Polyp  Segmentation, ISIC  2018, GLAnd  Segmentation (GLAS), and 2018  Data  Science, and the experimental results consistently demonstrate the superiorities of the proposed  DS-TransUNet. The main contributions of our work are as follows:

\begin{itemize}
\item[(1)] By incorporating the advantages of hierarchical Swin Transformer into both encoder and decoder, the proposed DS-TransUNet can effectively model long-range dependencies and multi-scale context connections during the process of down-sampling and up-sampling. To the best of our knowledge, this work is might be the first attempt to combine the  Swin Transformer with U-shaped architecture for automatic medical image segmentation. \par

\item[(2)] We introduce dual-branch Swin Transformer to extract multi-scale feature representations in the encoder, which enables the model to effectively capture coarse-fine-tuning features of different semantic scales, improving the quality of feature learning.\par

\item[(3)] The TIF module is able to establish effective global dependencies between coarse and fine-grained feature representations based on self-attention mechanism, which can guarantee the coarse-fine-tuning features of semantic consistency.\par

\item[(4)]Extensive experiments across four typical tasks for medical image segmentation show that the proposed DS-TransUNet consistently outperforms previous state-of-the-art methods especially in polyp segmentation task, which demonstrates the effectiveness of our method.
\end{itemize}

\section{Related Work}\label{related_work}
In this section, we first summarize the most typical CNN-based methods used in medical image segmentation, then we make a overview of the recent related works about vision transformers, especially in the filed of segmentation. Finally, we review the existing methods which perform multi-scale feature representations and compare these methods with our proposed method.\par
\subsection{Medical image segmentation based on CNNs}
Convolutional neural networks (CNNs), especially encoder-decoder based U-Net\cite{ronneberger2015u} and its variants have demonstrated superb performance in medical image segmentation, e.g., UNet++\cite{zhou2018unet++} designs a series of nested and dense skip pathways to reduce the semantic gap, Attention U-Net\cite{oktay2018attention} proposes a novel attention gate (AG) mechanism that enables the model to focus on targets of 
different shapes and sizes, Res-UNet\cite{xiao2018weighted} adds weighted attention mechanism to improve the performance of retinal vessel segmentation, DenseUNet\cite{li2018h} takes the advantages of dense connections and skip connection of U-Net, R2U-Net\cite{alom2018recurrent} combines the strengths of residual networks and U-Net to achieve better feature representation, KiU-Net\cite{valanarasu2020kiu} proposes a novel architecture utilizing both under-complete and over-complete features that makes improvement in segmenting small anatomical structures, DobuleU-Net\cite{jha2020doubleu} uses two U-Net in sequence and adopts Atrous Spatial
Pyramid Pooling (ASPP)\cite{chen2017rethinking}, UNet 3+\cite{huang2020unet} leverages deep supervisions and full-scale skip connections, and feed attention network (FANet)\cite{tomar2021fanet} unifies the previous epoch mask with the current epoch feature map during training. Note that all these methods are still based on CNNs.

\subsection{Vision Transformer}
Inspired by the success of Transformer\cite{vaswani2017attention} in various NLP tasks, more and more Transformer-based methods appear in CV tasks. Among the recent vision transformers, ViT\cite{dosovitskiy2020image} is the first attempt that proves pure Transformer-based architecture can achieve SOTA performance on image recognition when pre-training on large datasets such as ImageNet-22K and JFT-300M. DeiT\cite{touvron2020training} introduces data-efficient training strategies and knowledge distillation that allow ViT to perform well on smaller ImageNet-1K dataset. Swin Transformer\cite{liu2021swin} has linear computational complexity through proposed shifted window based self-attention and achieves SOTA performance in image recognition, dense prediction tasks such as object detection and semantic segmentation. Unlike most previous Transformer-based models, Swin Transformer is a hierarchical architecture which has the flexibility to be a general-purpose backbone network. SETR\cite{zheng2020rethinking} treats semantic segmentation as a sequence-to-sequence prediction task by using transformer as encoder. In medical image segmentation, TransUNet\cite{chen2021transunet} proves that Transformer can be used as powerful encoders for medical image segmentation. TransFuse\cite{zhang2021transfuse} is proposed to improve efficiency for global context modeling by fusing transformers and CNNs. Furthermore, to train the model effectively on medical images, MedT\cite{valanarasu2021medical} introduces Gated Axial-Attention based on Axial-DeepLab\cite{wang2020axial}. Inspired by these approaches, we propose a UNet-like architecture which applies Swin Transformer block to both encoder and decoder. It is our belief that a unified architecture across encoder and decoder based on Transformer could provide strong performance in medical image segmentation.

\subsection{Multi-Scale Tranformer}
Multi-scale feature representations based on CNNs are a classic concept in computer vision, and have shown to benefit various CV tasks\cite{cai2016unified}\cite{nah2017deep}\cite{chen2018big}\cite{cheng2020higherhrnet}. Especially, the classic feature pyramid networks (FPN)\cite{lin2017feature} has been widely adopted in object detection and semantic segmentation. However, such benefits have not been explored much in vision transformers. The close works include: CrossViT\cite{chen2021crossvit} proposes a dual-branch transformer and cross-attention for image classification. M2TR\cite{wang2021m2tr} introduces a multi-scale transformer that operates on different patch sizes of feature representations. MViT\cite{fan2021multiscale} provides a multi-scale pyramid of features inside the transformers. Motivated by the great potential of multi-scale vision transformers, we propose a dual-branch encoder which benefits from the hierarchical architecture of Swin Transformer. Moreover, we design a efficient module called Transformer Interactive Fusion (TIF) module to fuse the multi-scale feature representations.

\begin{figure}
\centering
\subfigure[] { \label{trans1}
\includegraphics[width=0.12\textwidth]{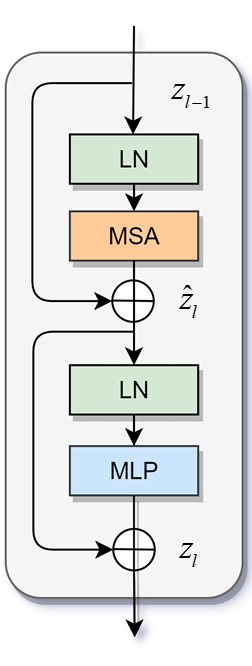}}\hspace{2.1mm}
\subfigure[] { \label{trans2}
\includegraphics[width=0.33\textwidth]{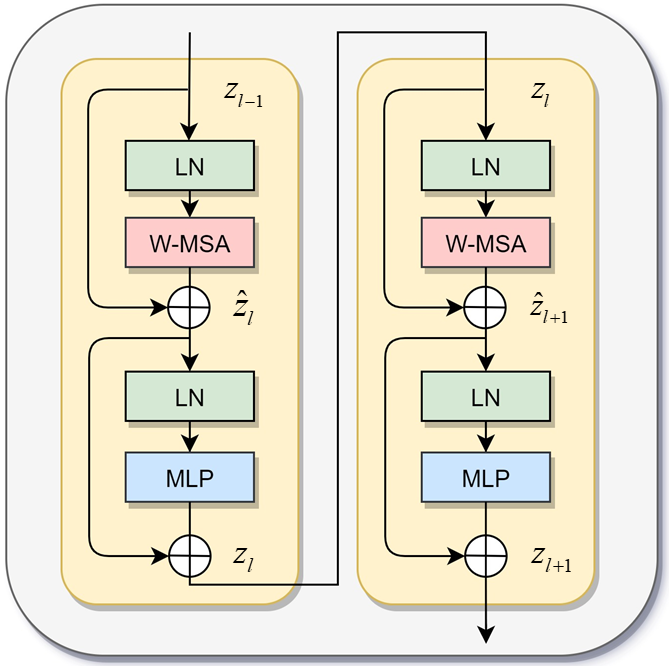}}
\caption{(a) The architecture of a standard Transformer block (notation presented with Eq. (\ref{equ1})); (b) The architecture of a Swin Transformer block (notation presented with Eq. (\ref{equ2}) and Eq. (\ref{equ3})).}
\end{figure}

\section{Method}\label{method}
\begin{figure*}
    \centering
    \includegraphics[width=0.8\textwidth]{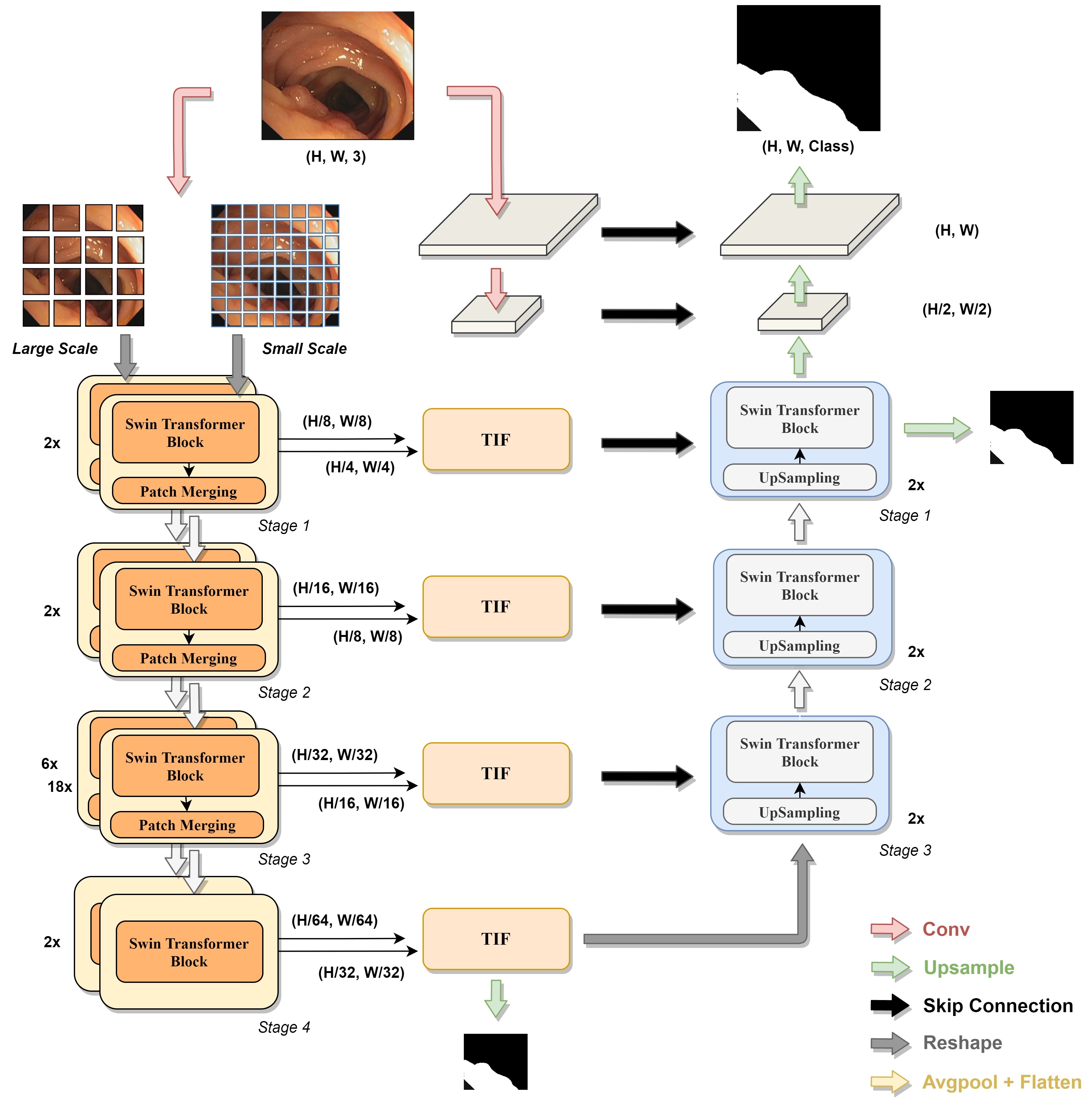}
    \caption{Illustration of the proposed Dual Swin Transformer U-Net (DS-TransUNet). Given an input medical image, we first split it into non-overlapping patches at two scales and feed them into the two branches of encoder separately, then the output feature representations of different scales will be fused by Transformer Inter-active Fusion (TIF) module. Finally, the fused features are restored to the same resolution as input image after the up-sampling process based on Swin Transformer block, hence obtaining the final mask predictions.}
    \label{fig2}
\end{figure*}

In this section, the overall structure of proposed DS-TransUNet is introduced in detail and illustrated in Fig. \ref{fig2}. We first introduce the standard Transformer and Swin Transformer adopted in DS-TransUNet, then we elaborate the encoder and decoder based on Swin Transformer block since our model is a U-shaped architecture. Finally, we show that our DS-TransUNet can benefit from the dual-branch encoder design and describe how multi-scale feature representations are effectively fused by Transformer Interactive Fusion (TIF) module.

\subsection{Swin Transformer block}
The standard Transformer encoder\cite{vaswani2017attention} is composed of a stack of \emph{L} identical blocks. As shown in \ref{trans1}, each block is consist of Multi-head Self Attention (MSA) and Multi Layer Perceptron (MLP). Besides, a LayerNorm (LN) layer is applied before each MSA module and each MLP, and a residual connection is applied after each module. Therefore the output $z_l$ of \emph{l}-layer in Transformer encoder can be expressed as:
\begin{equation}\label{equ1}\begin{aligned}
&\hat{\mathbf{z}}_{l}=\mathrm{MSA}\left(\mathrm{LN}\left(\mathbf{z}_{l-1}\right)\right)+\mathbf{z}_{l-1}, \\
&\mathbf{z}_{l}=\mathrm{MLP}\left(\mathrm{LN}\left(\hat{\mathbf{z}}_{l}\right)\right)+\hat{\mathbf{z}}_{l},\\
\end{aligned}
\end{equation}
\par
In the standard Transformer architecture , every token needs to be computed its relationships with all other tokens, where the computational complexity is quadratic equal to the number of tokens, making it unacceptable for many dense prediction and high-resolution image tasks. For efficient modeling, Swin Trasnformer \cite{liu2021swin} proposes Window based MSA (W-MSA) and Shifted Window based MSA (SW-MSA).\par
In W-MSA, the input feature will be divided into non-overlapping windows, and each window contains $M\times M$ patches (set to 7 by default). W-MSA will only conduct self-attention within local windows. As shown in Fig. \ref{trans2}, $\hat{z_l}$ and $z_l$ represent the outputs of W-MSA and MLP in $l^{th}$ layer , which are computed as:
\begin{equation}\label{equ2}\begin{aligned}
&\hat{\mathbf{z}}_{l}=\mathrm{W\mbox{-}MSA}\left(\mathrm{LN}\left(\mathbf{z}_{l-1}\right)\right)+\mathbf{z}_{l-1}, \\
&\mathbf{z}_{l}=\mathrm{MLP}\left(\mathrm{LN}\left(\hat{\mathbf{z}}_{l}\right)\right)+\hat{\mathbf{z}}_{l}, \\
\end{aligned}
\end{equation}
\par
The problem of W-MSA is the lack of effective information interaction between windows, in order to introduce cross-window interaction without additional computation, there exists a SW-MSA followed by the W-MSA.\par
The window configuration of SW-MSA is different from the previous W-MSA layer where it proposes an efficient batch processing method by cyclic-shifting to the upper-left. After this shift, a batch window may be consisted of multiple non-adjacent sub-windows in the feature map and keep the equal number of batch windows as regular partitioning at the same time. While conducting self-attention within local windows in both W-MSA and SW-MSA, the relative position bias is included in computing similarity.\par
With such shifted window partitioning mechanism, the outputs of SW-MSA and MLP module can be written as:
\begin{equation}\label{equ3}\begin{aligned}
\hat{\mathbf{z}}_{l+1}=\mathrm{SW\mbox{-}MSA}\left(\mathrm{LN}\left(\mathbf{z}_{l}\right)\right)+\mathbf{z}_{l}, \\
\mathbf{z}_{l+1}=\mathrm{MLP}\left(\mathrm{LN}\left(\hat{\mathbf{z}}_{l+1}\right)\right)+\hat{\mathbf{z}}_{l+1}, \\
\end{aligned}
\end{equation}

\subsection{Encoder}\label{encoder}
In the overall structure of our model, we refer to \cite{ronneberger2015u} using the U-shaped architecture. For encoder, the Swin Transformer\cite{liu2021swin} is used for feature extraction. As shown in Fig. \ref{fig2}, the input medical image will first be sliced into $\frac{H}{s}\times\frac{H}{s}$ non-overlapping patches, where \emph{s} is the patch size. Each patch is treated as a “token” and will be projected to dimension \emph{C} by linear embedding layer. Since the patches are obtained by convolution operation, no additional position information is needed here. 
These patch tokens are formally fed into Swin Transfomer, which contains four stages, and each stage holds a certain number of Swin Transformer blocks that include window multi-head self attention (W-MSA) and shifted window multi-head self attention (SW-MSA). To produce a hierarchical representation, the number of tokens will be reduced as the network gets deeper; 
in the first three stages, input features will go through patch merging layer to reduce the feature resolution and increase dimension after Swin Transformer blocks’ transformation. Specifically, the patch merging layer concatenates features of each group of $2\times2$ neighboring patches, and then applies a linear layer on the channel-dimensional concatenated features. This will reduce the number of tokens by $2\times2=4$, $2\times$ downsampling of resolution and increase the output dimension by 2. So the output resolutions of four stages are $\frac{H}{s}\times\frac{H}{s}$, $\frac{H}{2s}\times\frac{H}{2s}$, $\frac{H}{4s}\times\frac{H}{4s}$ and $\frac{H}{8s}\times\frac{H}{8s}$; and the dimensions are \emph{C}, 2\emph{C}, 4\emph{C} and 8\emph{C} respectively.

\subsection{Decoder}\label{decoder}
As shown in Fig. \ref{fig2}, the decoder mainly consists of three stages. Unlike the previous U-Net \cite{ronneberger2015u} and its variants, each stage of our model includes not only up-sampling (Nearest Upsampling) and skip connection, but also Swin Transfomer block. Specifically, the output of stage 4 in encoder is used as initial input of decoder. In each stage of decoder, the input features are up-sampled by 2, and then concatenated with the appropriate skip connection feature maps from encoder in the same stage. After that, the output is fed into Swin Transformer block. We choose this design since 1) it allows us to make full use of the features from encoder and up-sampling 2) it can build long-range dependencies and global context interaction in decoder to achieve better decoding performance. The impact of introducing Swin Transformer block in decoder will be discussed in section \ref{ablation_study}.\par
After the three stages above, we can get the output with resolution of $\frac{H}{4}\times\frac{H}{4}$. Using a $4\times$ upsampling operator directly will lost a lot of shallow features, so we down-sampling the input image by cascading two blocks to get the low level feature with resolution of $H\times W$ and $\frac{H}{2}\times\frac{H}{2}$, where each block consists a $3\times3$ convolutional layer, a group normalization layer and a ReLU layer successively. All these output features will be used to get the final mask predictions through skip connection.

\subsection{Multi-Scale Feature Representations}\label{multi-scale}
Although self-attention can effectively build long-range dependencies between patches, patch division ignores the pixel-level intrinsic structure features inside each patch, which will lead to the lose of shallow features such as edges and lines information. Moreover, ViT \cite{dosovitskiy2020image} can obtain better performance with fine-grained patch size. Taking these into account, and in order to improve the segmentation performance and enhance the robustness of our model, we employ multi-scale Swin Transformer for feature extraction.\par
Patches of different scales can complement each other in feature extraction; the large scale can better capture coarse-grained feature, while small patch can better obtain the fine-grained feature. Although the convolutional layer can introduce location information between patches implicitly, the information is lost at pixel level within each patch. In \cite{chen2021crossvit}, dual-branch Transfomrer can alleviate the above problems to a certain extent, and achieve better performance than ViT in image recognition. Motivated by this, we propose multi-scale Swin Transformer in encoder. More specifically, we use two independent branch with patch size of $s=4$ (primary) and $s=8$ (complementary) for feature extraction at different spatial levels. As result, the output with resolutions of $\frac{H}{4}\times\frac{H}{4}$,  $\frac{H}{8}\times\frac{H}{8}$, $\frac{H}{16}\times\frac{H}{16}$ and $\frac{H}{32}\times\frac{H}{32}$ can be obtained from small-scale branch, while output resolutions of large-scale are $\frac{H}{8}\times\frac{H}{8}$, $\frac{H}{16}\times\frac{H}{16}$, $\frac{H}{32}\times\frac{H}{32}$ and $\frac{H}{64}\times\frac{H}{64}$.

\begin{figure*}
    \centering
    \includegraphics[width=1.\textwidth]{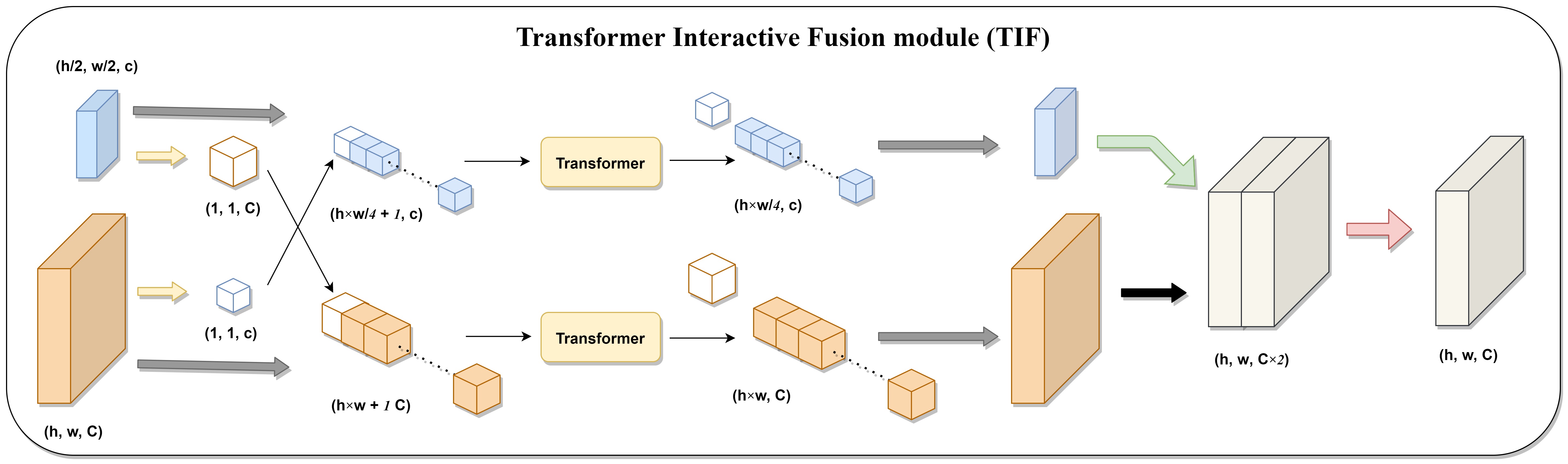}
    \caption{Illustration of Transformer Interactive Fusion module (TIF), which serves as the core component of DS-TransUNet in the multi-scale features fusion process.}
    \label{fig3}
\end{figure*}
\subsection{Transformer Interactive Fusion Module (TIF)}\label{TIF}
After obtaining the output features from dual-branch encoder, the remaining problem is how to fuse them since effective feature fusion is the core of multi-scale feature representations learning. A direct approach is to simply concatenate the multi-scale features and then perform convolution operation. However, such straightforward approach fails to capture the long-range dependencies and global context connection between features at different scales. Therefore, we propose a novel Transformer Interactive Fusion (TIF) module, which utilizes the MSA mechanism to enable efficient and effective interaction between multi-scale features. In particular, we select the standard Transformer block\cite{vaswani2017attention} instead of Swin Transformer block in TIF, mainly because the latter essentially operates on rectangle-based feature map, while in multi-scale features fusion module, we need to generate a token at specified size based on feature map of one branch, and then compute self-attention together with the token sequence reshaped by another branch. Moreover, we only need to perform monolayer self-attention operation twice at each stage, which means the computational complexity is acceptable.\par
As shown in Fig. \ref{fig3}, the proposed TIF can integrate features from two branches of different scales. In the following, we choose the small scale branch for specific analysis, and the same procedure is also applicable to large scale branch. \par
To be specific, for outputs of two branches from the same stage \emph{i} ($i=1, 2, 3, 4$) denoted as $F^i=[f^i_1, f^i_2,...,f^i_{h\times w}]\in\mathbb{R}^{C\times(h\times w)}$ (primary branch) and $G^i=[g^i_1, g^i_2,...,g^i_{\frac{h}{2}\times\frac{w}{2}}]\in\mathbb{R}^{c\times(\frac{h}{2}\times\frac{w}{2})}$ (complementary branch), respectively. Then we obtain the transformation output of $G^i$ by:
\begin{equation}\label{equ4}\begin{aligned}
\hat{\mathbf{g}}^{i}=\mathrm{Flatten}\left(\mathrm{Avgpool}\left({G}^{i}\right)\right), \\
\end{aligned}
\end{equation}
where $\hat{g}^i\in\mathbb{R}^{C\times 1}$, Avgpool is a 1 dimension average pooling layer, followed by flatten operation. The token $\hat{g}^i$ represents the global abstract information of $G^i$ to interact with $F^i$ at pixel level . Meanwhile, $F^i$ is concatenated with $\hat{g}^i$ into a sequence of $1+h\times w$ tokens, which is fed into Transformer layer for computing global self-attention:
\begin{equation}\label{equ5}\begin{aligned}
\hat{{F}}^{i}&=\mathrm{Transformer}\left([\hat{g}^i, f^i_1, f^i_2,...,f^i_{h\times w}]\right), \\
&=[\hat{f}^i_0, \hat{f}^i_1,...,\hat{f}^i_{h\times w}]\in\mathbb{R}^{C\times(1+h\times w)}\\
{F}^i_{out}&=[\hat{f}^i_1, \hat{f}^i_2,...,\hat{f}^i_{h\times w}]\in\mathbb{R}^{C\times(h\times w)} \\
\end{aligned}
\end{equation}
where \emph{Transformer} plays the same role as Eq. \ref{equ1} and ${F}^i_{out}$ as the final output of small scale branch in TIF. This approach introduces connections between each token in ${F}^i=[{f}^i_1, {f}^i_2,...,{f}^i_{h\times w}]\in\mathbb{R}^{C\times(h\times w)}$ and the whole $G^i$, so that fine-grained feature  can also obtain coarse-grained information from the large scale branch. Therefore, the TIF module can bring effective feature fusion of multi-scale branch which helps achieve better segmentation performance. The impact of TIF compared to ordinary multi-scale features fusion based on CNN will be discussed in section \ref{ablation_study}.\\

\section{Experiments}\label{experiments}
To evaluate the the learning and generalization ability of our Dual Swin Transformer U-Net (DS-TransUNet), we conduct experiments on four common medical image segmentation tasks with several publicly available datasets, and compare them with other SOTA methods. In this section we present the basic information about all the datasets briefly. Besides, we also describe the evaluation metrics and implementation details.

\subsection{Datasets}

\begin{figure*}
\centering
\subfigure[Qualitative results on Kvasir dataset (train-ing/testing split:880/120)] { \label{polyp1}
\includegraphics[width=0.99\textwidth]{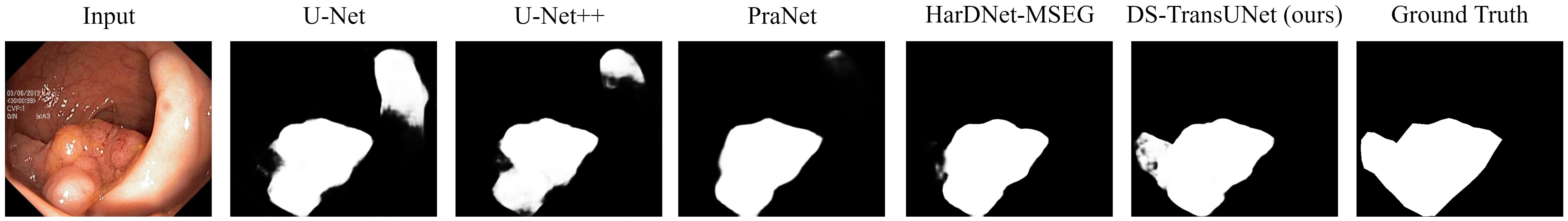}}
\subfigure[Qualitative results on CVC-ClinicDB dataset (training/testing split:550/62)] { \label{polyp2}
\includegraphics[width=0.99\textwidth]{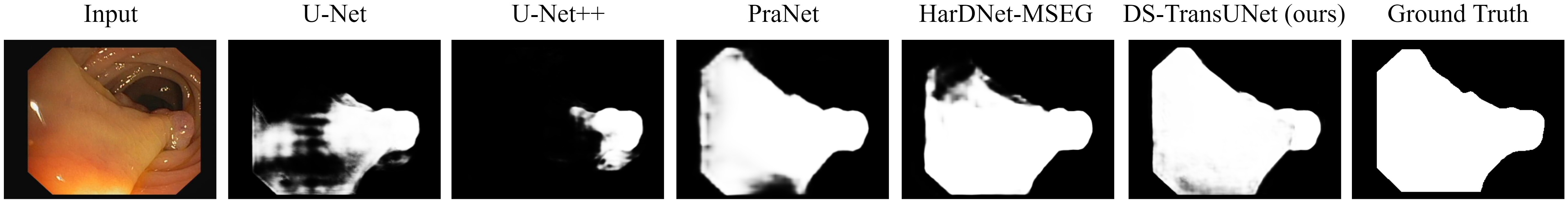}}
\subfigure[Qualitative results on five polyp segmentation datasets] { \label{polyp3}
\includegraphics[width=0.99\textwidth]{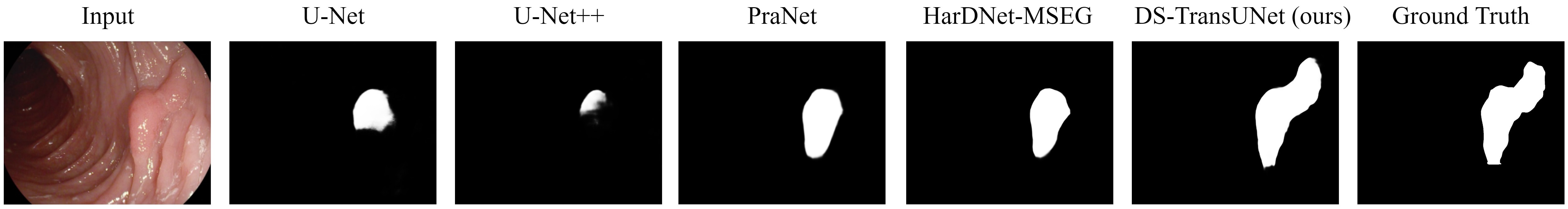}}
\caption{Qualitative results on polyp segmentation task of DS-TransUNet compared to other models. Our model shows better learning and generalization ability, which leads to higher-quality segmentation performance.}

\end{figure*}

\textbf{Polyp Segmentation:} For polyp segmentation task, we select five public polyp datasets including Kvasir \cite{jha2020kvasir}, CVC-ColonDB (ColonDB) \cite{bernal2015wm}, EndoScene \cite{tajbakhsh2015automated}, ETIS \cite{vazquez2017benchmark}, and CVC-ClinicDB (ClinicDB) \cite{silva2014toward}. The split and training settings of these datasets are different in \cite{fan2020pranet}, \cite{jha2020doubleu} and \cite{jha2021real}, so we conduct experiments according to these three articles respectively.\par
\begin{itemize}
\item In \cite{jha2021real}, only Kvasir is used, which contains 880 images for training and 220 images for testing. For this split, we resize each image to a resolution of $512\times512$.\par
\item According to \cite{jha2020doubleu}, we only use ClinicDB during experiment, of which 550 images are used for training while 62 for testing. Besides, all the images used are resized to $384\times384$.\par
\item As for \cite{fan2020pranet}, the training sets consist 900 images in Kvasir and 550 images in ClinicDB, while the testing sets contain five datasets, which are Kvasir with 100 images, ClinicDB with 62 images, ColonDB with 380 images, EndoScene with 60 images and ETIS with 196 images. Since the resolutions of images in datasets are not uniform, we resize them to $384\times384$.\par
\end{itemize}\par
\textbf{ISIC 2018 Dataset:} The dataset comes from ISIC-2018 challenge \cite{codella2019skin}\cite{tschandl2018ham10000} and is useful for skin lesion analysis. It includes 2596 images and their corresponding annotations, which are resized to $256\times256$. The images are randomly split into 2076 images for training and 520 images for testing. This process is repeated five times and the average is taken as result.\par
\textbf{GLAS Dataset:} GLAnd Segmentation (GLAS) datatset is from 2015 challenge on gland segmentation in histology images, which provides images of Haematoxylin and Eosin (H\&E) stained slides. It contains 165 images which are split into 85 images for training and 80 for testing according to \cite{valanarasu2020kiu}. Besides, images are resized into $128\times128$.\par
\textbf{2018 Data Science Bowl:} The dataset is from 2018 Data Science Bowl challenge \cite{caicedo2019nucleus} and used to find the nuclei in divergent images, including 670 images in total. We use the same settings as \cite{fan2020pranet}, 80\% of dataset for training, 10\% for validation, and 10\% for testing. Moreover, all the images are resized into $256\times256$.

\subsection{Evaluation Metrics}
To compare SOTA methods with our proposed DS-TransUNet, the standard evaluation metrics that we use include mean Dice Coefficient (mDice) (a.k.a. F1), mean Intersection over Union (mIoU), precision and recall, which are associated with four values i.e., true-positive (TP), true-negative (TN), false-positive (FP), and false-negative (FN).\par
\begin{equation}\label{equ8}\begin{aligned}
{mDice}&=\displaystyle{\frac{2\times{TP}}{2\times{TP}+FP+FN}}, \\
{mIoU}&=\displaystyle{\frac{{TP}}{{TP}+FP+FN}}, \\
{Precision}&=\displaystyle{\frac{{TP}}{TP+FP}}, \\
{Recall}&=\displaystyle{\frac{{TP}}{TP+FN}}, \\
\end{aligned}
\end{equation}

\subsection{Implementation Details}
Multi-scale training strategy is used in all experiments instead of data augmentation. The loss functions used are weighted IoU loss $\mathcal{L}^{W}_{IoU}$ and binary cross-entropy loss $\mathcal{L}^{W}_{BCE}$. Inspired by \cite{fan2020pranet}, we find deep supervision helps the model training by additionally supervising the output \emph{S2} of stage 4 in encoder and \emph{S3} of stage 1 in decoder, which means the final loss function $\mathcal{L}_{total}$ can be written as:
\begin{equation}\label{equ6}\begin{aligned}
&\mathrm{\mathcal{L}}_{total}=\alpha\mathrm{\mathcal{L}}(G, S_1)+\beta\mathrm{\mathcal{L}}(G, S_2)+\gamma\mathrm{\mathcal{L}}(G,S_3),\\
&\mathrm{\mathcal{L}}=\mathrm{\mathcal{L}}^{W}_{IoU}+\mathrm{\mathcal{L}}^{W}_{BCE},
\end{aligned}
\end{equation}
where \emph{G} is the groundturth in training sample and $\alpha,\beta,\gamma$ are hyperparameters which are set to 0.6, 0.2, 0.2 empirically. We train our model with SGD optimizer with momentum 0.9, weight decay 1e-4 and learning rate equals to 0.01.\par
All models are trained for 100 epochs. Moreover, early stopping and Cosine Annealing schedule are also used. All models are built using PyTorch framework and trained on a NVIDIA RTX 3090 GPU. Our model is provided in two variants: the base version (DS-TransUNet-B) uses Swin-Base\cite{liu2021swin} as primary scale branch (small scale branch) for encoder, while the large version (DS-TransUNet-L) uses Swin-Large\cite{liu2021swin}. Both the two version use Swin-Tiny\cite{liu2021swin} as complementary scale branch (large scale branch) for encoder. All these sizes of Swin Transformer use pretrained weights released from \cite{liu2021swin}. The detailed parameters of model are summarized in Table \ref{tab1}, where Layer Number and Head Number mean the number of Swin Transformer block and head self-attention in each stage respectively. Moreover, Window Size represents the size of non-overlapping windows divided in W-MSA, and Swin-Decoder refers to the Swin Transformer block used in decoder.

\begin{table*}\centering\caption{DETAILS OF SWIN TRANSFORMER MODEL VARIANTS.}
\renewcommand\arraystretch{1.5}
\begin{tabular}{l|ccccc}
\bottomrule
\textbf{Methods}      & \textbf{Hidden Size \emph{C}} & \textbf{MLP Size \emph{D}} & \textbf{Layer Number} & \textbf{Head Number} & \textbf{Window Size} \\ \hline\hline
Swin-Tiny   & 96    &384     & [2, 2, 6, 2]        & [3, 6, 12, 24]   & 7                    \\
Swin-Base    & 128   &512    & [2, 2, 18, 2]       & [4 ,8, 16, 32]     & 7                    \\
Swin-Large   & 192   &768     & [2, 2, 18, 2]       & [6, 12, 24, 48]    & 7                    \\
Swin-Decoder & 128   &512    & [2, 2, 2]           & [8, 4, 2]          & 7  \\ \bottomrule               
\end{tabular}
\label{tab1}
\end{table*} 
 
\section{Result}

\begin{figure*}
\centering
\subfigure[Quantitative results on ISIC 2018 dataset] { \label{isic}
\includegraphics[width=0.99\textwidth]{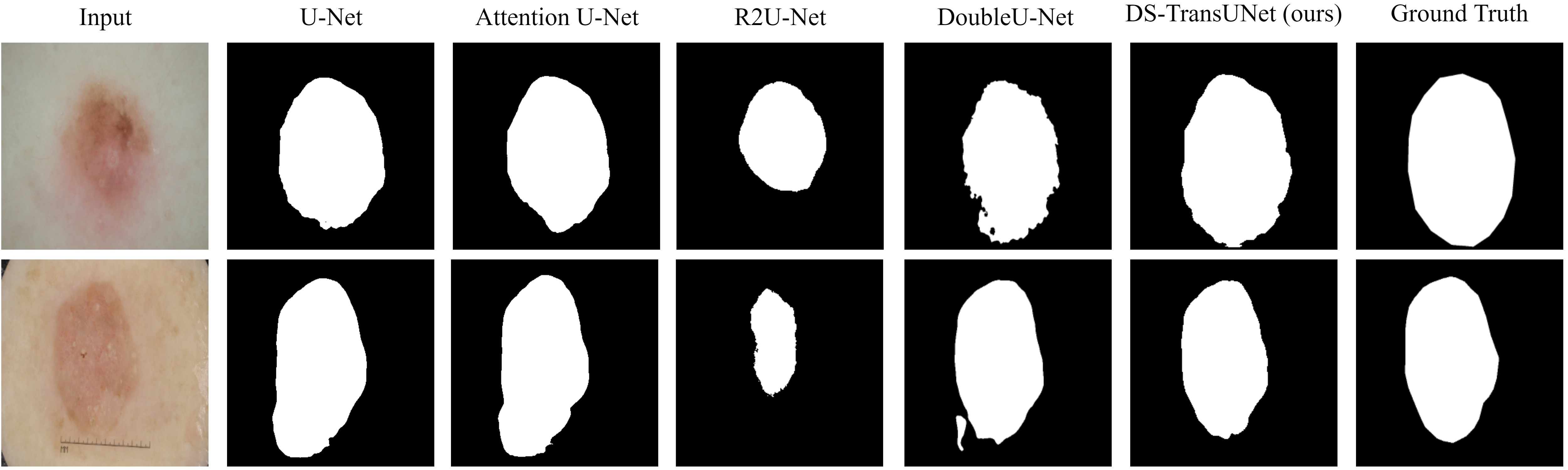}}
\subfigure[Quantitative results on GLAS dataset] { \label{glas}
\includegraphics[width=0.99\textwidth]{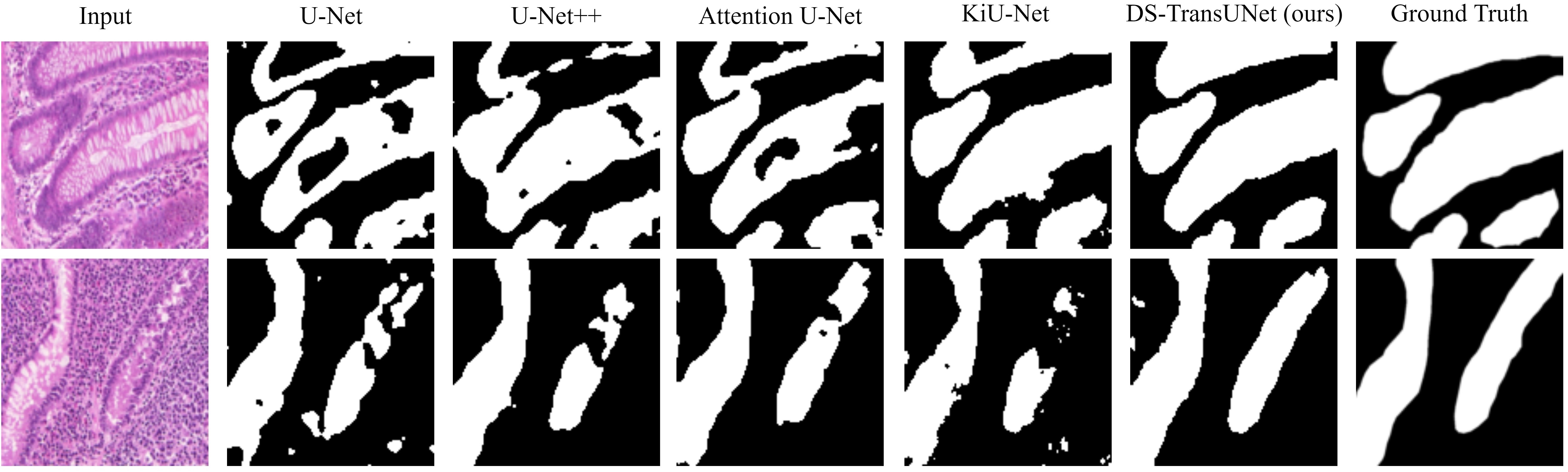}}
\subfigure[Quantitative results on 2018 Data Science Bowl dataset] { \label{bowl}
\includegraphics[width=0.99\textwidth]{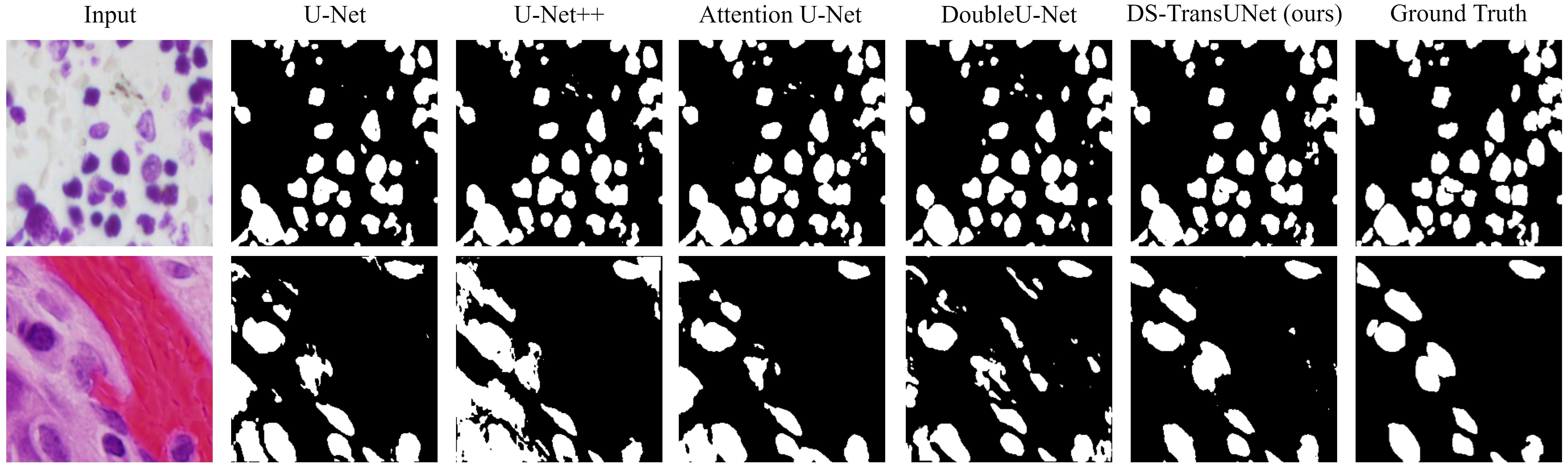}}
\caption{Qualitative results of DS-TransUNet on three medical image segmentation tasks compared to other models.}
\vspace{0.2in}
\end{figure*}

In this section, we conduct experiments to compare our proposed model with SOTA methods on four segmentations tasks. Besides, we also present the experimental results and visualize some qualitative results to evaluate the learning and generalization ability of our DS-TransUNet. Finally, we also perform ablation study on polyp segmentation task to analyze the effect of each proposed technique uesd in DS-TransUNet.
\begin{table}\centering
\caption{QUANTITATIVE RESULTS ON KVASIR DATASET (TRAINING/TESTING SPLIT:880/120). FOR EACH COLUMN, THE BEST RESULTS ARE HIGHLIGHTED IN \textbf{BOLD}.}
\renewcommand\arraystretch{1.5}
\begin{tabular}{l|cccc}
\bottomrule
\textbf{Method}                          & mDice & mIou  & Recall & Precision\\ \hline\hline
U-Net\cite{ronneberger2015u}                   & 0.597          & 0.471       & 0.617    & 0.672   \\
Res-UNet\cite{xiao2018weighted}              & 0.690          & 0.572       & 0.725    & 0.745   \\
ResUNet++\cite{jha2019resunet++}               & 0.714          & 0.613       & 0.742    & 0.784   \\
DoubleU-Net\cite{jha2020doubleu}             & 0.813          & 0.733       & 0.840    & 0.861   \\
FCN8\cite{long2015fully}                    & 0.831          & 0.737       & 0.835    & 0.882   \\
PSPNet\cite{zhao2017pyramid}                  & 0.841          & 0.744       & 0.836    & 0.890   \\
HRNet\cite{wang2020deep}                   & 0.845          & 0.759       & 0.859    & 0.878   \\
DeepLabv3+\cite{chen2018encoder} & 0.864          & 0.786       & 0.859    & 0.906   \\
FANet\cite{tomar2021fanet}                     & 0.880          & 0.810       & 0.906    & 0.901   \\
HarDNet-MSEG\cite{huang2021hardnet}              & 0.904          & 0.848       & 0.923    & 0.907   \\ \hline
DS-TransUNet-B (ours)                & 0.911          & 0.856       & 0.935    & 0.914  \\
\textbf{DS-TransUNet-L (ours)}      & \textbf{0.913} & \textbf{0.859} & \textbf{0.936} & \textbf{0.916}\\ \bottomrule
\end{tabular}
\label{tab2}
\end{table}

\begin{table}\centering\caption{QUANTITATIVE RESULTS ON CVC-ClINICDB DATASET (TRAINING/TESTING SPLIT:550/62). FOR EACH COLUMN, THE BEST RESULTS ARE HIGHLIGHTED IN \textbf{BOLD}.}
\renewcommand\arraystretch{1.5}
\begin{tabular}{l|cccc}
\bottomrule
\textbf{Method}      & F1     & mIoU   & Recall & Precision \\ \hline\hline
SFA\cite{fang2019selective} &0.7000 & 0.6070 & - & - \\
ResUNet-mod\cite{zhang2018road}               & 0.7788          & 0.4545 & 0.6683               & 0.8877                  \\
UNet++\cite{zhou2018unet++}               & 0.7940          & 0.7290 & -               & -                  \\
ResUNet++\cite{jha2020doubleu}   & 0.7955          & 0.7962          & 0.7022          & 0.8785    \\
U-Net\cite{ronneberger2015u}                & 0.8230          & 0.7550          & -               & -                  \\
PraNet\cite{fan2020pranet} & 0.8990 & 0.8490 & - & - \\
DoubleU-Net\cite{jha2020doubleu} & 0.9239 & 0.8611 & 0.8457 & \textbf{0.9592} \\
FANet\cite{tomar2021fanet}         & 0.9355          & 0.8937          & 0.9339          & 0.9401             \\ \hline
DS-TransUNet-B (ours)  & 0.9350          & 0.8845          & 0.9464       & 0.9306             \\
\textbf{DS-TransUNet-L (ours)} & \textbf{0.9422} & \textbf{0.8939}          & \textbf{0.9500} & 0.9369             \\ \bottomrule
\end{tabular}
\label{tab8}
\end{table}

\subsection{Comparison with State-of-the-art Methods}

\textbf{Results on Polyp Segmentation:} Our quantitative results on polyp segmentation task achieve SOTA performance compared to other models, which is present in Table \ref{tab2}, \ref{tab8} and \ref{tab3}. Next, we analyze the quantitative results on the three kinds of data splits.\par

In \cite{jha2021real}, only Kvasi dataset \cite{jha2020kvasir} is used and the evaluation metrics used include mDice, mIoU, recall and precision. From Table \ref{tab2}, we can see that not only DS-TransUNet-L, but also DS-TransUNet-B outperforms the previous SOTA HarDNet-MSE\cite{huang2021hardnet} on all metrics. Specifically, DS-TransUNet-L achieves a mDice of 0.913, mIoU of 0.859, recall of 0.936 and precision of 0.916 with an improvement of 0.9\%, 1.1\%, 1.3\% and 0.9\%. As visualized in Fig. \ref{polyp1}, our DS-TransUNet achieve the best segmentation performance among all models, especially for fuzzy polyps at the edge of image, which are often missed-out in colonoscopy because their color and structure are similar to the surrounding intestinal tissue.\par

According to \cite{jha2020doubleu}, only CVC-ClinicDB dataset is used during experiment. Table \ref{tab8} shows that our proposed DS-TransUNet-L achieves SOTA results on almost all metrics (F1, mIoU, and recall). Specifically, DS-TransUNet-L outperforms the previous SOTA FANet\cite{tomar2021fanet} by an improvement of 0.67\%, 0.02\% and 1.6\% in terms of F1, mIoU and recall, while produces a comparable precision score compared to DoubleU-Net\cite{jha2020doubleu}. From Fig. \ref{polyp2}, we can see that polyps with large area can also be accurately segmented. Moreover, the higher recall score also shows that our DS-TransUNet is more clinically useful.\par

Referring to\cite{fan2020pranet}, where the training sets are consist of Kvasir and  CVC-ClinicDB, and testing sets additionally include three unseen datasets. As for quantitative evaluation, we use mDice and mIoU following\cite{fan2020pranet}. Our proposed model achieves SOTA performance on all five challenging dataset, which is the only model that produces over 0.92 and 0.93 mDice on Kvasir. In particular, DS-TransUNet-B and DS-TransUNet-L both outperform the latest TransFuse\cite{zhang2021transfuse} on the two in-domain datasets. As for unseen datasets (ColonDB, EndoSene and ETIS), our DS-TransUNet also greatly exceeds all SOTA methods by a large margin. Specifically, DS-TransUNet-B achieves better performance on all datasets except EndoScene, while DS-TransUNet-L even yields the top performance on all five datasets. In general, our proposed method outperforms SOTA methods with mDice improvement of 1.7\%, 0.4\%, 2.5\%, 0.7\% and 3.5\%, and achieves about 1.9\% improvement in terms of the average mDice score compared with TransFuse, which shows the advantages and strong learning ability of our proposed model. The qualitative segmentation performance in Fig. \ref{polyp3} also shows the great generalization ability of DS-TransUNet.\par

\begin{table}\centering\caption{QUANTITATIVE RESULTS ON ISIC 2018 DATASET. FOR EACH COLUMN, THE BEST RESULTS ARE HIGHLIGHTED IN \textbf{BOLD}.}
\renewcommand\arraystretch{1.5}
\begin{tabular}{l|cccc}
\bottomrule
\textbf{Method}      & F1    & mIoU   & Recall & Precision \\ \hline\hline
U-Net\cite{ronneberger2015u}                & 0.6740          & 0.5490          & 0.7080          & -                  \\
Attention U-Net\cite{oktay2018attention}      & 0.6650          & 0.5660          & 0.7170          & -                  \\
R2U-Net\cite{alom2018recurrent}              & 0.6790          & 0.5810          & 0.7920          & -                  \\
Attention R2U-Net\cite{alom2018recurrent}    & 0.6910          & 0.5920          & 0.7260          & -                  \\
BCDU-Net (d=3)\cite{azad2019bi}       & 0.8510          & -               & 0.7850          & -                  \\
FANet\cite{tomar2021fanet}          & 0.8731          & 0.8023          & 0.8650          & 0.9235             \\
DoubleU-Net\cite{jha2020doubleu}    & 0.8962          & 0.8212          & 0.8780          & \textbf{0.9459}    \\ \hline
DS-TransUNet-B (ours)  & 0.9101          & 0.8481          & 0.9108          & 0.9337             \\
\textbf{DS-TransUNet-L (ours)} & \textbf{0.9132} & \textbf{0.8523} & \textbf{0.9217} & 0.9271             \\ \bottomrule
\end{tabular}
\label{tab5}
\end{table}

\textbf{Results on ISIC 2018 Dataset:} For 2018 ISIC dataset, the metrics used are F1 Score, mIoU, recall and precision. Table \ref{tab5} presents the specific results, where our proposed model achieves better segmentation performances than SOTA DoubleU-Net \cite{jha2020doubleu} and the latest FANet \cite{tomar2021fanet}. DS-TransUNet-L produces 0.9132 on F1, 0.8523 on mIoU and recall of 0.9217 with an improvement of 1.70\%, 3.11\% and 4.37\% compared with SOTA method, respectively. Although DoubleU-Net outperforms in terms of precision, our proposed model produces the best overall performances on four metrics. As shown in Fig. \ref{isic}, the qualitative results qualitatively manifest that our proposed method can not only accurately predict the location and boundary of skin lesion, but also better distinguish it from normal skin.\par

\begin{table*}\centering\caption{QUANTITATIVE RESULTS ON FIVE POLYP SEGMENTATION DATASETS COMPARED TO PREVIOUS SOTA METHODS. FOR EACH COLUMN, THE BEST RESULTS ARE HIGHLIGHTED IN \textbf{BOLD}.}
\renewcommand\arraystretch{1.5}
\begin{tabular}{l|llllllllll|ll}
\bottomrule
\multirow{2}{*}{}           & \multicolumn{2}{c}{Kvasir} & \multicolumn{2}{c}{ClinicDB} & \multicolumn{2}{c}{ColonDB} & \multicolumn{2}{c}{EndoScene} & \multicolumn{2}{c}{ETIS} & \multicolumn{2}{|c}{Average} \\
                            & mDice   & mIoU    & mDice   & mIoU     & mDice   & mIoU    & mDice   & mIoU     & mDice   & mIoU   & mDice   & mIoU \\ \hline\hline

U-Net\cite{ronneberger2015u}                                                 & 0.818            & 0.746            & 0.823             & 0.755             & 0.512             & 0.444            & 0.398              & 0.335             & 0.710           & 0.626           & 0.652             & 0.581            \\
U-Net++\cite{zhou2018unet++}                                               & 0.821            & 0.743            & 0.794             & 0.729             & 0.483             & 0.410            & 0.401              & 0.344             & 0.707           & 0.624           & 0.641             & 0.570            \\
PraNet\cite{fan2020pranet}                                                & 0.898            & 0.840            & 0.899             & 0.849             & 0.709             & 0.640            & 0.871              & 0.797             & 0.628           & 0.567           & 0.800             & 0.739            \\
HarDNet-MSEG\cite{huang2021hardnet}                                          & 0.912            & 0.857            & 0.932             & 0.882             & 0.731             & 0.660            & 0.887              & 0.821             & 0.677           & 0.613           & 0.828             & 0.767            \\
TransFuse-S\cite{zhang2021transfuse}                                           & 0.918            & 0.868            & 0.918             & 0.868             & 0.773             & 0.696            & 0.902              & 0.833             & 0.733           & 0.659           & 0.849             & 0.785            \\
TransFuse-L\cite{zhang2021transfuse}                                           & 0.918            & 0.868            & 0.934             & 0.886             & 0.744             & 0.676            & 0.904              & 0.838             & 0.737           & 0.661           & 0.847             & 0.786            \\ \hline
\textbf{DS-TransUNet-B (ours)}                                   & 0.934            & 0.888            & \textbf{0.938}    & \textbf{0.891}    & 0.798             & 0.717            & 0.882              & 0.810             & \textbf{0.772}  & \textbf{0.698}  & 0.865             & 0.801            \\
\textbf{DS-TransUNet-L (ours)}                                  & \textbf{0.935}   & \textbf{0.889}   & 0.936             & 0.887             & \textbf{0.798}    & \textbf{0.722}   & \textbf{0.911}     & \textbf{0.846}    & 0.761           & 0.687           & \textbf{0.868}    & \textbf{0.806}  \\ \bottomrule
\end{tabular}
\label{tab3}
\end{table*}

\begin{table}\centering\caption{QUANTITATIVE RESULTS ON THE GLAS DATASET. FOR EACH COLUMN, THE BEST RESULTS ARE HIGHLIGHTED IN \textbf{BOLD}.}
\renewcommand\arraystretch{1.5}
\begin{tabular}{l|cc}
\bottomrule
\textbf{Method}            & mDice & mIoU \\ \hline\hline
Seg-Net\cite{badrinarayanan2017segnet}                   & 78.61                     & 65.96                  \\ 
U-Net\cite{ronneberger2015u}                      & 79.76                     & 67.63                  \\
MedT\cite{valanarasu2021medical}                 & 81.02                     & 69.61\\
UNet++\cite{zhou2018unet++}                      & 81.13                     & 69.61                  \\
Attention UNet\cite{oktay2018attention}                      & 81.59                     & 70.06                  \\
KiU-Net\cite{valanarasu2020kiu} & 83.25                     & 72.78                  \\ \hline
DS-TransUNet-B (ours)        & 86.54                     & 77.36                  \\
\textbf{DS-TransUNet-L (ours) }       & \textbf{87.19}            & \textbf{78.45}         \\ \bottomrule
\end{tabular}
\label{tab7}
\end{table}

\textbf{Results on GLAS Dataset:} Our quantitative results on the GLAS dataset are shown in Table \ref{tab7}. Comparing with the leading SOTA method KiU-Net \cite{valanarasu2020kiu}, our proposed methods DS-TransUNet-B and DS-TransUNet-L both outperform KiU-Net on both mDice and mIoU. Especially DS-TransUNet-L achieves a 3.94\% improvement in terms of mDice and 5.67\% of mIoU over SOTA method. GLAS is a dataset with only 85 training samples, and we can achieve SOTA performance only by using multi-scale training, which effectively proves that our proposed method can produce high-quality segmentation performance even on a small-scale datasets. Besides, we also present the visualization of generated mask images in Fig. \ref{glas}, which demonstrates that our model is able to better distinguish the gland itself from the surrounding tissue, and bring excellent gland segmentation performance.\par

\begin{table}\centering\caption{QUANTITATIVE RESULTS ON THE 2018 DATA SCIENCE BOWL. FOR EACH COLUMN, THE BEST RESULTS ARE HIGHLIGHTED IN \textbf{BOLD}.}
\renewcommand\arraystretch{1.5}
\begin{tabular}{l|cccc}
\bottomrule
\textbf{Method}      & F1     & mIoU   & Recall & Precision \\ \hline\hline
U-Net\cite{ronneberger2015u}                & 0.7573          & 0.9103          & -               & -                  \\
UNet++\cite{zhou2018unet++}               & 0.8974          & \textbf{0.9255} & -               & -                  \\
Attention UNet\cite{oktay2018attention}               & 0.9083          & 0.9103 & -               & 0.9161                  \\
DoubleU-Net\cite{jha2020doubleu}   & 0.9133          & 0.8407          & 0.6407          & \textbf{0.9496}    \\
FANet\cite{tomar2021fanet}         & 0.9176          & 0.8569          & 0.9222          & 0.9194             \\ \hline
\textbf{DS-TransUNet-B (ours)}  & 0.9200          & 0.8589          & \textbf{0.9427}          & 0.9054             \\
\textbf{DS-TransUNet-L (ours)} & \textbf{0.9219} & 0.8612          & 0.9378 & 0.9124             \\ \bottomrule
\end{tabular}
\label{tab6}
\end{table}

\textbf{Results on 2018 Data Science Bowl:} For 2018 data science bowl challenge, we compare our result with the SOTA models. Table \ref{tab6} shows that DS-TransUNet-L achieves a F1 of 0.9219, mIoU of 0.8612 and recall of 0.9378, which are 0.43\%, 0.43\% and 1.56\% higher than the best performing FANet\cite{tomar2021fanet}. Besides, DS-TransUNet-B can yield the highest recall score of 0.9427. In general, although UNet++ and DoubleU-Net still keep the SOTA performance in mIoU (0.9255) and precision (0.9496) respectively, our proposed model achieves the best balance among the four metrics compared to the other SOTA methods. From the qualitative results in Fig. \ref{bowl}, we can observe that our DS-TransUNet can better capture the presence of cell nuclei and bring better segmentation prediction.

\begin{table*}[htp]\centering\caption{BLATION STUDY ON POLYP SEGMENTATION TASK. FOR EACH COLUMN, THE BEST RESULTS ARE HIGHLIGHTED IN \textbf{BOLD}.}
\renewcommand\arraystretch{1.5}
\begin{tabular}{l|llllllllll|ll}
\bottomrule
\multirow{2}{*}{}           & \multicolumn{2}{c}{Kvasir} & \multicolumn{2}{c}{ClinicDB} & \multicolumn{2}{c}{ColonDB} & \multicolumn{2}{c}{EndoScene} & \multicolumn{2}{c}{ETIS} & \multicolumn{2}{|c}{Average} \\
                            & mDice   & mIoU    & mDice   & mIoU     & mDice   & mIoU    & mDice   & mIoU     & mDice   & mIoU   & mDice   & mIoU \\ \hline\hline
TransFuse-S         & 0.918            & 0.868            & 0.918             & 0.868             & 0.773             & 0.696            & 0.902              & 0.833             & 0.733           & 0.659          & 0.849 & 0.785 \\
TransFuse-L         & 0.918            & 0.868            & 0.934             & 0.886             & 0.744             & 0.676            & 0.904              & 0.838             & 0.737           & 0.661    & 0.847 & 0.786       \\ \hline
Base model (Base)  & 0.919            & 0.863            & 0.915             & 0.861             & 0.747             & 0.657            & 0.878              & 0.804             & 0.722           & 0.635        & 0.836 & 0.764   \\
Base model (Large) & 0.922            & 0.867            & 0.914             & 0.861             & 0.786             & 0.698            & 0.884              & 0.812             & 0.735           & 0.652         & 0.848 & 0.778  \\ \hline
Swin U-Net (Base)   & 0.920            & 0.868            & 0.914             & 0.862             & 0.758             & 0.674            & 0.884              & 0.811             & 0.711           & 0.636        & 0.837 & 0.770   \\
Swin U-Net (large)   & 0.926            & 0.876            & 0.923             & 0.875             & 0.791             & 0.709            & 0.889              & 0.816             & 0.734           & 0.650         & 0.853 & 0.785  \\ \hline
Swin Decoder (Base)      & 0.927            & 0.877            & 0.936             & 0.889             & 0.785             & 0.697            & 0.886              & 0.813             & 0.741           & 0.666         & 0.855 & 0.788  \\
Swin Decoder (Large)   & 0.929            & 0.879            & 0.929             & 0.880             & 0.798             & 0.717            & 0.904              & 0.836             & 0.759           & 0.677         & 0.864 & 0.798  \\ \hline
Multi-Scale SD (Base)   & 0.931            & 0.882            & 0.927             & 0.878             & 0.784             & 0.704            & 0.864              & 0.789             & 0.716           & 0.632       & 0.844 & 0.777    \\
Multi-Scale SD (Large)  & 0.927            & 0.876            & 0.928             & 0.877             & 0.786             & 0.707            & 0.862              & 0.785             & 0.737           & 0.655       & 0.848 & 0.780    \\ \hline
\textbf{DS-TransUNet-B}  & 0.934            & 0.888            & \textbf{0.938}    & \textbf{0.891}    & 0.798             & 0.717            & 0.882              & 0.810             & \textbf{0.772}  & \textbf{0.698}  & 0.865 & 0.801 \\
\textbf{DS-TransUNet-L} & \textbf{0.935}   & \textbf{0.889}   & 0.936             & 0.887             & \textbf{0.798}    & \textbf{0.722}   & \textbf{0.911}     & \textbf{0.846}    & 0.761           & 0.687    & \textbf{0.868} & \textbf{0.806}\\ \bottomrule      
\end{tabular}
\label{tab4}
\end{table*}

\subsection{Ablation study}\label{ablation_study}
In order to evaluate the ability of Swin Transformer in medical image segmentation and the influence of various factors on our proposed model, we further conduct ablation studies on four variants of our DS-TransUNet. The datasets we select are based on polyp segmentation task, which can verify the learning and generalization ability of models.
\begin{itemize}
\item\textbf{Base model}, which directly processes the final output of Swin Transformer by a progressive upsampling strategy. Specifically, the output in stage 4 of Swin Transformer is up-sampled by cascading  three blocks, where each block consists convolution layers and $2\times$ upsampling operations. After that, we obtain the output with resolution of $\frac{H}{4}\times\frac{H}{4}$, and then perform the same processing as described in \ref{decoder} to make the final pixel-level predictions.
\item\textbf{Swin U-Net}, which is based on U-shape architecture. It uses Swin Transformer as encoder, while keeps the same structure in decoder as \cite{ronneberger2015u}, which only utilizes convolution layer, $2\times$ upsampling and skip connection.
\item\textbf{Swin Decoder}, which is based on Swin U-Net, further adding Swin Transformer block for long-range dependencies modeling after each up-sampling process. The specific Swin Transformer block parameters used in decoder are shown in Table \ref{tab1}.
\item\textbf{Multi-Scale SD}, whose full name is Multi-Scale Swin Decoder, leverages dula-branch Swin Transformer for feature extraction in encoder based on Swin Encoder. Compared with DS-TransUNet, it utilizes convolution layer for multi-scale feature representations fusion instead of TIF.
\end{itemize}\par
Table \ref{tab4} presents the experimental results of four variants of DS-TransUNet on polyp segmentation task, in terms of both mean Dice and mean IoU. Moreover, we select the latest TransFuse \cite{zhang2021transfuse} as baseline.\par

\textbf{Effect of Swin Transformer:} Swin Transformer block is the core component of our proposed method, which computes representation with W-MSA and SW-MSA, and surpasses the previous SOTA methods in multiple CV tasks. To explore the feature extraction ability of Swin Transformer in medical image segmentation task, we compare Base model with the previous SOTA TransFuse. In Table \ref{tab4} we can see that Swin Transformer achieves satisfied segmentation performance as encoder. Although it is not as good as TransFuse in overall performance, it still produces close and comparable results. Especially Base model (Large), shows an improvement of 0.4\% and 1.3\% in Kvasir and ColonDB in terms of mDice respectively compared to the best results of TransFuse.\par

\textbf{Effect of Swin Transformer block in decoder:} In order to explore the influence of Swin Transformer in decoder, we conduct the experiments of two specially designed models based on Swin Transformer as encoder: Swin U-Net and Swin Decoder. The specific results shown in Table \ref{tab4} indicate that the U-shaped encoder-decoder based architecture can effectively improve the segmentation performance. Especially Swin U-Net (Large), has achieved 0.4\% improvement in terms of the average mean Dice score compared to TransFuse.\par
By simply adding Swin Transformer block after each up-sampling in Swin U-Net, Swin Decoder can easily build long-range dependencies and global context connection in decoder. As shown in Table \ref{tab4}, we can see that Swin Decoder already achieves better performance than the latest TransFuse on all five challenging datasets with an improvement of 1.5\% in terms of the average mDice score, which means that Swin Decoder has better learning and generalization ability than previous SOTA methods. Specially, the best results of Swin Decoder outperform TransFuse with mDice improvement of 1.1\%, 0.2\%, 2.1\% and 2.2\% in all dataset except EndoScene. Therefore, the decoder design based on Swin Transformer block can effectively improve the segmentation performance.\par

\textbf{Effect of multi-scale feature representations and TIF:} Multi-Scale SD adds another Swin Transformer branch in encoder, and simply fuses the multi-scale features through convolution operation. Such a straightforward approach does not bring performance improvements. The experimental results shown in Table \ref{tab4} indicates that despite the additional encoder branch is added which brings more granular information, it fails to achieve better performance compared to Swin Encoder with single branch. This is mainly because common convolution layer can not effectively fuse multi-scale feature, but makes the model more difficult to converge\par
By adding TIF module to Multi-scale SD, we can get the final proposed DS-TransUNet, which yields the best performance among all variants. To evaluate the effectiveness of the proposed TIF module, we compare DS-TransUNet with Swin Decoder in Table \ref{tab4}. It can be observed that the best results of DS-TransUNet achieve mDice improvements of 0.6\%, 0.2\%, 0.7\%, 1.3\% on Kvasir, ClinicDB, EndoScene and ETIS compared to Swin Decoder with single branch, while show a 0.5\% improvement on ColonDB in terms of mIoU. In general, TIF allows more efficient interaction between features of different scales, which brings more effective feature representations fusion of multi-scale branches and helps achieve better segmentation performance.\par

\section{Conclusion}
In this work we present the Dual Swin Transformer U-Net (DS-TransUNet), a U-shaped encoder-decoder based framework for medical image segmentation. Our DS-TransUNet is based on the hierarchical Swin Transformer. Not only the encoder, we also innovatively add Swin Transformer block in decoder. Moreover, we introduce dual-branch Swin Transformer in encoder to extract multi-scale feature representations. We further propose a novel Transformer Interactive Fusion (TIF) module to build long-range dependencies between features of different scales through self-attention mechanism, thus effectively fusing the multi-scale features from encoder. Extensive experiments on four medical image segmentation tasks show that our DS-TransUNet significantly outperforms other state-of-the-art methods especially in polyp segmentation task. In the future, our work will focus on designing more lightweight Transformer-based models and better learning the pixel-level intrinsic structural features generated by the patch division in vision transformers.


%





\ifCLASSOPTIONcaptionsoff
  \newpage
\fi



%
\bibliographystyle{IEEEtran}
\bibliography{bare_jrnl.bbl}

%









\end{document}